\documentclass{article}

\usepackage{spconf}
\usepackage{amsmath,amssymb}
\usepackage{graphicx}
\usepackage{booktabs}
\usepackage{multirow}
\usepackage{multicol}
\usepackage{float}
\usepackage{makecell}
\usepackage{hyperref} % 一定放最后
\title{RemoteDet-Mamba: A Hybrid Mamba-CNN Network for Multi-modal Object Detection in Remote Sensing Images}
\name{ \fontsize{9.5}{10}\selectfont Kejun Ren$^{1}$ \qquad Xin Wu$^{1}$ \qquad Lianming Xu$^{2}$ \qquad Li Wang$^{1}$
\thanks{
This work was supported in part by the National Natural Science Foundation of China under Grants 62101045 and 62171054, and in part by the Natural Science Foundation of Beijing Municipality under Grant L222041, in part by the Fundamental Research Funds for the Central Universities under Grants No. 24820232023YQTD01, 2024RC06, and 2023RC96, in part by the Double First-Class Interdisciplinary Team Project Funds 2023SYLTD06. 
{Emails:\{kejun.ren, xin.wu, xulianming, liwang\}@bupt.edu.cn} (Corresponding author: \emph{Li Wang}.)}}

\address{$^{1}$ School of Computer Science, Beijing University of Posts and Telecommunications, Beijing, China \\
         $^{2}$ School of Electronic Engineering, Beijing University of Posts and Telecommunications, Beijing, China}

\begin{document}
\ninept
\maketitle
\begin{abstract}
Unmanned Aerial Vehicle (UAV) remote sensing, with its advantages of rapid information acquisition and low cost, has been widely applied in scenarios such as emergency response. However, due to the long imaging distance and complex imaging mechanisms, targets in remote sensing images often face challenges such as small object size, dense distribution, and low inter-class discriminability. To address these issues, this paper proposes a multi-modal remote sensing object detection network called RemoteDet-Mamba, which is based on a patch-level four-direction selective scanning fusion strategy. This method simultaneously learns unimodal local features and fuses cross-modal patch-level global semantic information, thereby enhancing the distinguishability of small objects and improving inter-class discrimination. Furthermore, the designed lightweight fusion mechanism effectively decouples densely packed targets while reducing computational complexity. Experimental results on the DroneVehicle dataset demonstrate that RemoteDet-Mamba achieves superior detection performance compared to current mainstream methods, while maintaining low parameter count and computational overhead, showing promising potential for practical applications.
\end{abstract}
\begin{keywords}
Unmanned aerial vehicle, remote sensing, multimodal, object detection, mamba.
\end{keywords}
\section{Introduction}
\label{sec:intro}

Unmanned aerial vehicles (UAVs) have become a flexible and cost-effective complement to traditional satellite remote sensing platforms, particularly in tasks such as environmental monitoring \cite{wu2021deep}, disaster response, and urban surveillance \cite{wu2020vehicle}. As shown in \cite{wu2021convolutional}, equipped with diverse sensors such as infrared (IR) and visible light (RGB) cameras, UAVs enable all-weather, high-resolution monitoring of ground objects. However, effective multimodal object detection remains challenging due to significant discrepancies between modalities, as well as the need to balance detection accuracy and computational efficiency.

While CNNs and Transformers have been widely adopted for multimodal feature fusion \cite{hong2023cross,li2024casformer, yuan2022translation}, their respective limitations—restricted receptive fields and high computational overhead—hinder their deployment on UAV platforms. Recent efforts combining CNNs with Transformer modules (e.g., C\textsuperscript{2}Former \cite{yuan2024c} have partially mitigated these issues but still inherit substantial complexity. Lately, the Mamba architecture \cite{gu2023mamba} has emerged as a promising alternative to traditional attention mechanisms, enabling global context modeling with linear computational complexity. Building upon its efficient state-space modeling capabilities, several Mamba-based variants, such as Vision Mamba \cite{zhu2024vision} and VMamba \cite{liu2024vmambavisualstatespace}, have demonstrated competitive performance across a wide range of vision and remote sensing tasks \cite{xiao2024frequency,chen2024rsmamba,ma2024rs}. Its recent extensions to multimodal learning have also shown encouraging results. For example, Fusion-Mamba \cite{dong2024fusion} introduces a hidden state space for cross-modal interaction, effectively aligning heterogeneous features and improving the consistency of fused representations. Sigma \cite{wan2024sigma} leverages a Siamese Mamba architecture for semantic segmentation, selectively integrating critical modality-specific information. DMM \cite{zhou2025dmm} further explores multimodal object detection by incorporating stacked Mamba modules for cross-modal feature extraction and hierarchical fusion.

\begin{figure}[tp]
  \centering
  \includegraphics[width=0.45\textwidth]{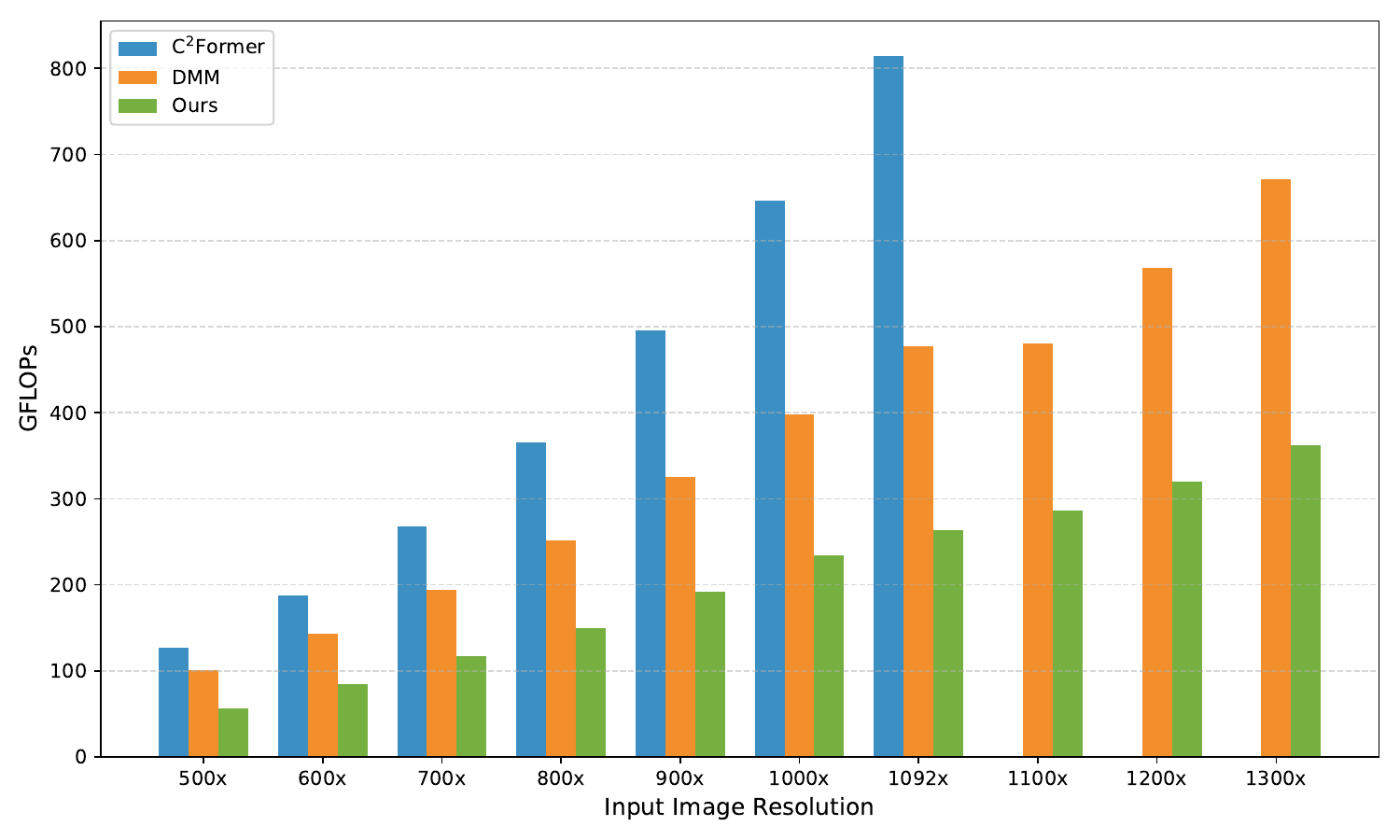}
  \caption{GFLOPS comparison with varying input sizes (H = W); C\textsuperscript{2}Former omitted beyond 1100×1100 due to quadratic memory growth.}
  \label{fig1}
\end{figure}

\begin{figure*}[ht]
  \centering
  \includegraphics[width=0.73\textwidth]{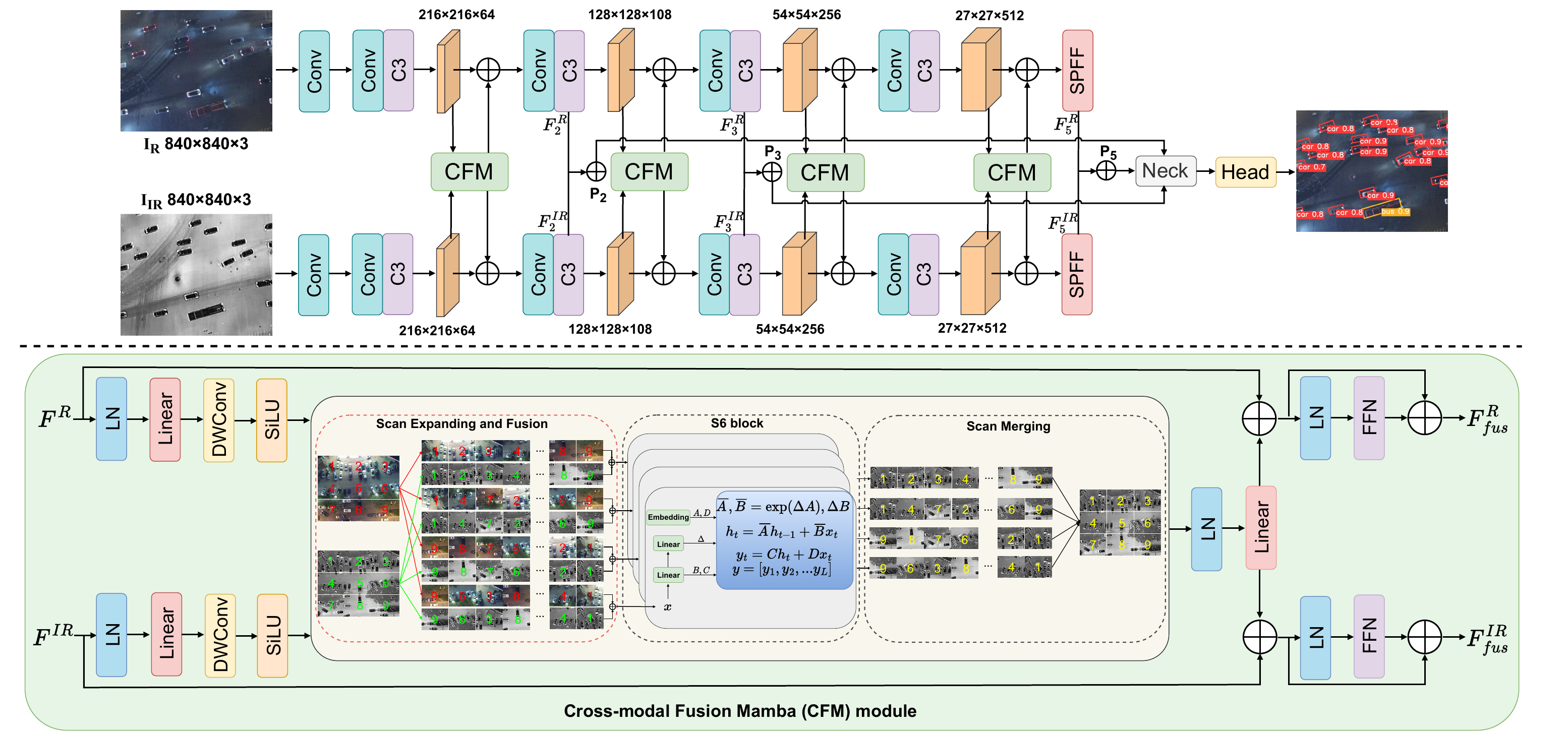}
  \caption{The architecture of the RemoteDet-Mamba framework. The top portion is the outline of the RemoteDet-Mamba. The bottom section provides a detailed view of our CFM module.}
  \label{fig.1}
\end{figure*}

Despite their effectiveness, these models often rely on deep and complex fusion pipelines or multiple sequential Mamba modules, resulting in substantial computational overhead, particularly at high input resolutions. As illustrated in Fig. \ref{fig1}, the FLOPs of existing methods scale sharply with image size, making them less suitable for real-time or resource-constrained UAV-based applications. To address this issue, we focus on the patch-level linear scanning mechanism within the only Mamba module and pose the following hypothesis: given that remote sensing targets are generally small in size and densely distributed, is this mechanism inherently well-suited for lightweight modality fusion in remote sensing object detection?

Based on this insight, we propose RemoteDet-Mamba, a novel framework tailored for multimodal UAV object detection in remote sensing imagery. The framework comprises a Siamese CNN encoder for multi-scale local feature extraction and a Cross-modal Fusion Mamba (CFM) module built upon Mamba’s Selective Scanning 2D mechanism (SS2D). The CFM module conducts four-directional linear scanning of the multi-scale features at the patch level, effectively decoupling densely distributed small targets and enabling selective, global feature fusion. This design achieves linear time complexity while capturing long-range dependencies, thus improving the discrimination of small objects and inter-class differentiation. Unlike existing approaches, our method avoids deep stacking of Mamba architecture and is particularly suited for small and densely distributed targets in UAV remote sensing images. The main contributions of this paper are as follows: 

\begin{itemize}
\item We propose RemoteDet-Mamba, a lightweight and efficient framework for multimodal UAV-based object detection in remote sensing imagery, integrating local feature extraction and global cross-modal fusion within a unified design. RemoteDet-Mamba avoids deep stacking of Mamba architecture and maintains strong cross-modal alignment and discriminative capability for small and densely distributed targets.

\item We design a novel CFM module, which utilizes Mamba's SS2D mechanism to perform four-directional linear scanning across patch-level features, enabling efficient global context modeling with linear time complexity.
\end{itemize}

\section{PROPOSED METHOD}
\label{sec:method}
Fig.\ref{fig.1} illustrates the architecture of the proposed RemoteDet-Mamba framework for multimodal UAV object detection, which comprises a Siamese CNN encoder and CFM module. Specifically, the Siamese CNN encoder extracts multi-scale features from each modality (e.g., RGB and TIR), enabling rich local representation learning. The CFM module, situated between the two modality-specific branches \cite{li2024learning}, performs deep patch-level global fusion via directional scanning, effectively aligning and integrating complementary information across modalities. The final fused feature is generated by aggregating the outputs from the CFM module together with the modality-specific features from both the visible and infrared branches.

\subsection{Modality-specific Features Learning}\label{AA}
Conventional multimodal networks often ignore the heterogeneous imaging mechanisms across modalities, which can lead to misaligned feature representations and degraded fusion performance. To address this, we adopt a Siamese CNN encoder architecture consisting of two structurally identical but independently parameterized branches, each dedicated to a specific modality (e.g., RGB or TIR). This design allows the network to preserve modality-specific features while avoiding premature feature entanglement.

Given two input images from different modalities, denoted as $I_s$ where $s \in \{R, IR\}$, the Siamese encoder extracts hierarchical feature maps through the following multi-scale convolutional operations:
\begin{equation}
F_i^s = \begin{cases} 
\text{Conv}(I_s), & \quad i = 0 \\ 
\text{Conv}(\text{C3}(F_{i-1}^s)), &  \quad i = 1, 2, 3, 4 \\
\text{SPFF}(F_{i-1}^s), &  \quad i = 5 ,
\end{cases}
\end{equation}
where $F_i^s$ denotes the $i$th-layer features of the $s$th modality. The specific details of $\text{C3}(\cdot)$ and $\text{SPFF}(\cdot)$ refer to YOLOv5s. $F_i^R$ and $F_i^{IR}$ represent the multi-scale features of two different modalities, respectively. 
\begin{equation}
P_i = F_i^R+F_i^{IR}, \quad i = 2,3,5. \\ 
\end{equation}

To construct multi-scale multimodal joint representations, modality-specific features from selected layers ($i = 2,3,5$) are element-wise summed to form preliminary joint features, which serve as input to the detection head. The Neck refines and integrates the multi-scale features $P_i$, which are then fed into the Head to produce the final remote sensing detection results.

\subsection{Cross-modal Fusion Mamba (CFM) module}
Unlike conventional fusion schemes relying on stacked attention modules, the CFM employs a single-layer four-directional SS2D scanning mechanism, which allows for long-range dependency modeling with linear time complexity. As shown in the dashed line in Fig.\ref{fig.1}, the CMF module receives multi-scale features $F^R$ and $F^{IR}$ from the Siamese CNN encoders introduced in Section~\ref{AA}. 

% The section below the dashed line in Fig.\ref{fig.1} illustrates the fusion process of the CFM module. Firstly, a set of multi-scale features $F^1$ and $F^{2}$ generated by Section \ref{AA} are sent to a LayerNorm layer to normalize the input features.
To begin, the modality-specific features $F^i$ are first normalized and projected into a unified latent space using LayerNorm (LN) and linear projection:

\begin{equation}
\overline F^i = \text{Linear}(\text{LN}(F^i)), \quad i \in \{R, IR\},%F^i = \{F^{1}, F^{2}\}
\end{equation}
where $\text{Linear}(\cdot)$ represents a linear projection layer. This step ensures that modality-specific statistics are normalized and embedded into a shared representation space, facilitating effective cross-modal fusion. %\textcolor{red}{These projected features are then processed by depthwise convolution to enhance intra-channel interactions.}

The resulting projected features $\overline{F}^R$ and $\overline{F}^{IR}$ are then further processed using depthwise convolution $\text{DWConv}(\cdot)$ to enhance intra-channel interactions.

\begin{equation}
f^i = \text{SiLU}(\text{DWConv}(\overline F^i)), \overline F^i  \in \{\overline F^{R}, \overline F^{IR}\},
\end{equation}
where $\text{SiLU}(\cdot)$ refers to the SiLU activation function. The Mamba architecture is based on a state space model and employs an input-driven selective scanning mechanism, which dynamically adjusts the internal state update process based on the input. This mechanism can be used to jointly model sequence features from different modalities, thereby facilitating information fusion between modalities. Inspired by VMamba \cite{liu2024vmambavisualstatespace}, the features $f^R$ and $f^{IR}$ are flattened into patch sequences along four scanning directions ($i=1,2,3,4$). This multi-directional flattening effectively decouples spatially adjacent but semantically unrelated small targets by dispersing them across different patch scan sequences, thereby reducing feature interference during fusion. 
\begin{equation}
\begin{split}
&f_i^1 = \text{flatten}_i(f^R),\\
&f_i^{\text{2}} = \text{flatten}_i(f^{\text{IR}}), \quad i=1,2,3,4\\
&f_i^{\text{FUS}} = f_i^R + f_i^{\text{IR}},
\end{split}
\end{equation}
where $\text{flatten}(\cdot)$ denotes the scanning operation along the i-th direction. Each patch fused sequence $f_i^{\text{FUS}}$ is then passed through a direction-specific S6 block for feature modeling, producing four outputs denoted as $y_1$, $y_2$, $y_3$, and $y_4$, respectively.
\begin{equation}
y_i = \text{S6}_i(f_i^{\text{FUS}}), \quad i=1,2,3,4,
\end{equation}
where $S6_i$ denotes the i-th S6 block. The outputs of the S6 blocks are unfolded and recombined to generate new feature maps, denoted as $Y^{\text{FUS}}$. These fused features are then projected back to the size of the original input feature space.
\begin{equation}
 Y^{\text{FUS}}=\sum_{i=1}^{4} \text{unflatten}_i(y_i),
\end{equation}
where $\text{unflatten}(\cdot)$ signifies the operation of reconstructing a one-dimensional fused sequence along the i-th direction into a two-dimensional feature map. The result $Y^{\text{FUS}}$, along with the original inputs from the two modalities, is then processed through residual connections to yield the complementary features $\hat{F}^R$ and $\hat{F}^{IR}$.
\begin{equation}
    \hat{F}^i = F^i + \text{Linear}(\text{LN}(Y^{\text{FUS}}),\quad i\in\{R, IR\}.
\end{equation}

Finally, to further enhance the discriminative capability of the fused features, each modality-specific output $\hat{F}^i$ is refined using a layer normalization followed by a feedforward network (FFN). %This lightweight refinement pipeline, when coupled with SS2D-based directional fusion, enables RemoteDet-Mamba to maintain high efficiency while enhancing feature discrimination for remote sensing objects.

\begin{equation}  
    F_{\text{fus}}^i = \hat{F}^i + \text{FFN}(\text{LN}(\hat{F}^i)), \quad i \in \{R, \text{IR}\} ,
\end{equation}  
\begin{equation}  
    \text{FFN}(\text{LN}(\hat{F}^i)) = \text{GELU}(\text{LN}(\hat{F}^i)W_1 + b_1)W_2 + b_2 ,
\end{equation}
where Gaussian Error Linear Units (GELU) is the nonlinear activation function. The parameters $W_1$, $W_2$, $b_1$, and $b_2$ represent the weight matrices and bias vectors, respectively, for the linear transformations within the FFN.

\subsection{Loss Function}
The total loss function $L_{total}$ of the proposed detection framework is composed of four parts: boundary regression loss $L_{box}$, confidence loss $L_{obj}$, classification loss $L_{cls}$, and angle classification loss $L_{theta}$. It is defined as follows:
\begin{equation}
    L_{total} = \alpha L_{box} + \beta L_{obj} + \gamma L_{cls} + \delta L_{theta},
\end{equation}
where $\alpha$,$\beta$,$\gamma$, and $\delta$ are set to 0.05, 1.0, 0.5, and 0.5, respectively. The Complete Intersection over Union (CIoU) was selected as the loss function $L_{box}$. For $L_{cls}$, we used Smooth Binary Cross Entropy \cite{mannor2005cross} (Smooth BCE) to enhance numerical stability. The same Smooth BCE was applied to compute the confidence loss $L_{obj}$, with CIoU added between horizontal edges to accelerate training. For the angular loss $L_{theta}$, we transform the regression problem into a classification problem through the use of CSL\cite{yang2020arbitrary}, calculated as:
\begin{equation}
CSL(x) = \begin{cases}
g(x), & \theta - r < x < \theta + r \\
0, & \text{otherwise},
\end{cases}
\end{equation}
where $g(x)$ represents a Gaussian window function characterized by periodicity, symmetry, monotonicity, and finiteness, and $r$ denotes the radius of the window:
\begin{equation}
    g(\text{angle}) = e^{-\text{angle}^2 / 2r^2}, \quad -0 \leq \text{angle} < 180^\circ.
\end{equation}

\section{EXPERIMENTS AND DISCUSSIONS}
\label{sec:experiment}

\subsection{Experimental Conditions}
\textbf{1) Data Description:} To evaluate the performance of the proposed RemoteDet-Mamba framework, we conduct both quantitative and qualitative experiments on the DroneVehicle dataset \cite{sun2022drone}. DroneVehicle is a large-scale benchmark specifically designed for RGB–infrared vehicle detection and counting tasks in UAV-based remote sensing scenarios. 

\textbf{2) Experimental Setup:} In the experiment, CSPDarkNet53 is utilized as the backbone network for the Siamese CNN, and data augmentation is implemented through random color transformations and image reconstruction. Model parameters are optimized using stochastic gradient descent (SGD), with an initial learning rate set to 1e-2, subsequently reduced to 2e-3. The proposed RemoteDet-Mamba is implemented in the PyTorch framework on a single NVIDIA GeForce RTX 3090, with Ubuntu 18.04 and CUDA version 12.0.

\subsection{Experiments and Discussions}

\textbf{Ablation analysis:} TABLE \ref{tbl:table1} confirms the superiority of our CFM module over additive and bidirectional scanning fusion. To handle the label mismatch between the two modalities, we use the larger ground truth box for training and testing, referred to as Fusion GT. Notably, the CFM outperforms bidirectional fusion by 0.7\% in mAP@0.5, indicating that the SS2D-based four-directional scanning more effectively captures global contextual dependencies. Unlike previous methods that perform local or pairwise fusion, CFM achieves more comprehensive feature interaction across modalities while maintaining low computational cost.

\begin{table}[ht]
    \centering
    \caption{Ablation analysis on the DroneVehicle dataset}
    \label{tbl:table1}
    \renewcommand\arraystretch{1.6}
    \setlength{\tabcolsep}{5pt} % Adjust column spacing to reduce table width
    \renewcommand{\arraystretch}{1.0}
    \scriptsize
    \begin{tabular}{c|cccc|cc}
        \Xhline{0.6pt}
        \multirow{2}{*}{\makecell{GT \\ Form}} & \multicolumn{4}{c|}{Fusion Strategy} & \multirow{2}{*}{\makecell{mAP@0.5\\(\%)}} & \multirow{2}{*}{\makecell{mAP@0.5:0.95\\(\%)}} \\ \cline{2-5}
         & No fusion & Add & Bid-scanning & CFM & &  \\
        \Xhline{0.6pt}        
         \multirow{4}{*}{\makecell{RGB \\ GT}} & \checkmark &  &  &  & 65.7 & 40.3 \\
         &  & \checkmark &  &  & 80.7 & 57.5 \\
         &  &  & \checkmark & & 80.8 & 58.2  \\
         & & & & \checkmark & 81.1 & 58.3  \\
        \Xhline{0.6pt}        
         \multirow{4}{*}{\makecell{Fusion \\ GT}} & \checkmark(RGB) & & & & 67.7 & 40.5 \\
         & \checkmark(TIR) &  &  & & 76.1 & 53.2 \\
         &  & \checkmark &  &  & 78.3 & 55.8  \\
         &  &  & \checkmark &  & 78.3 & 56.4  \\
         &  &  &  & \checkmark & 78.7 & 56.1  \\
        \Xhline{0.6pt}        
         \multirow{4}{*}{\makecell{TIR \\ GT}} & \checkmark &  &  &  &69.4 & 45.9 \\
         &  & \checkmark & & & 80.8 & 57.7 \\
         &  &  & \checkmark & & 81.1 & 58.5  \\
         &  &  &  & \checkmark & \textbf{81.8} & \textbf{58.9}  \\
        \Xhline{0.6pt}
    \end{tabular} 
    \vspace{-0.1mm}
   % \caption*{\tiny \textbf{Note:}No fusion refers to single modality.}
\end{table}

\begin{figure}[!t]
  \centering
  \includegraphics[width=0.33\textwidth]{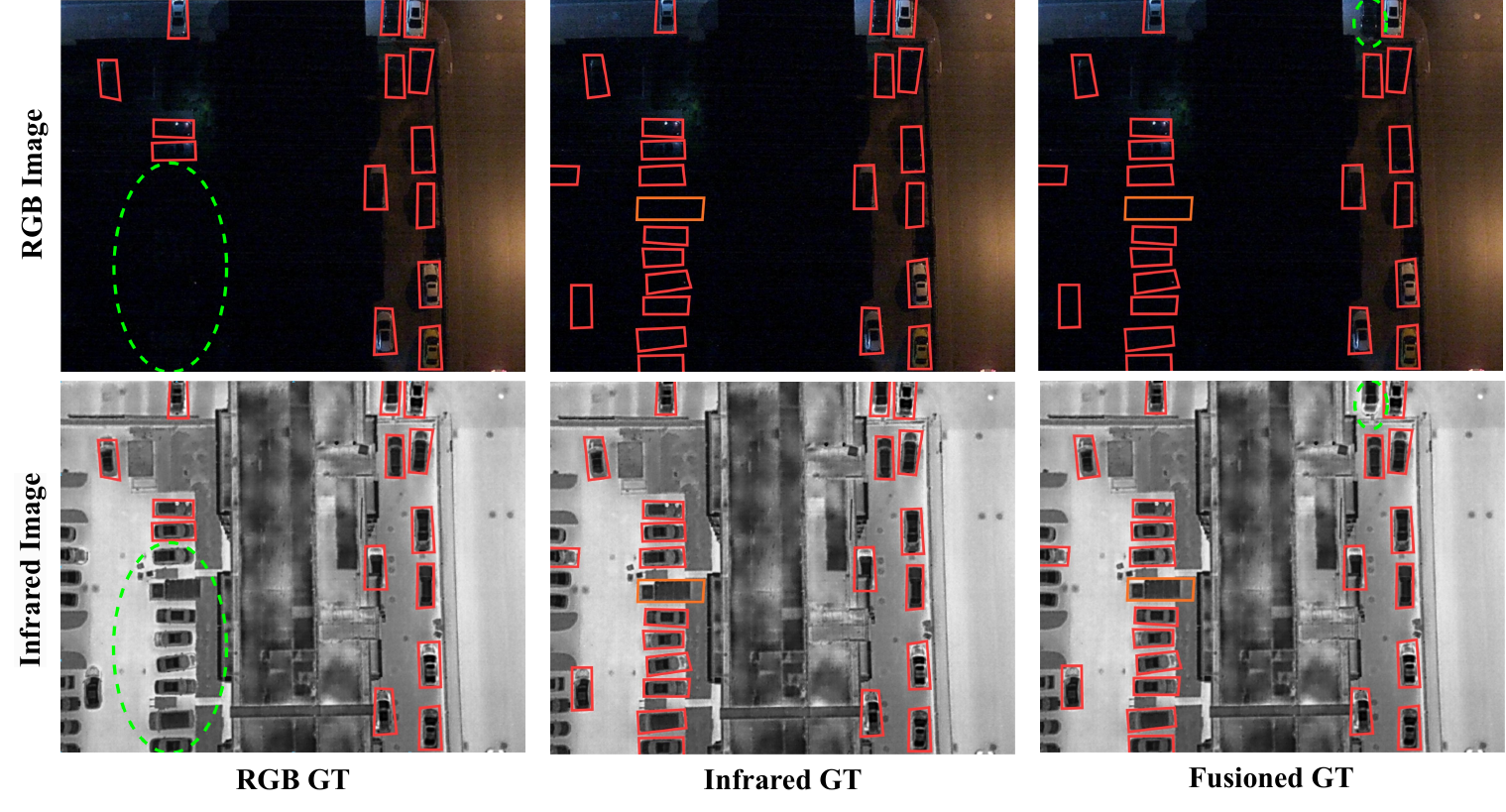}
  \caption{The visual ground truth boxes under different GT forms.}
  \label{fig.2}
\end{figure}

% Fig.\ref{fig.2} shows the ground truth boxes under different GT forms for training and testing. It shows that the thermal infrared ground truth boxes provide more accurate object labeling across varying lighting conditions than visible light and fused ground truth boxes. 
Fig. \ref{fig.2} shows the ground truth boxes under different GT forms for training and testing. RGB GT and TIR GT provide modality-specific supervision, with TIR GT offering more robust labeling under varying illumination. The fused GT is further introduced to enforce spatial consistency across modalities, mitigating cross-modal annotation misalignment. Similarly, TABLE \ref{tbl:table1} presents the quantitative results for different ground truth scenarios. When using only TIR or RGB images as inputs, the mean Average Precision (mAP) values are 0.657 and 0.694, respectively. With multi-modal inputs, employing addition fusion, bidirectional scanning fusion, and the proposed fusion strategy, the mAP values achieved are $0.808$, $0.811$, and $0.818$, respectively. This indicates that the proposed fusion method improves performance by 12.4\% over the thermal infrared unimodal approach, demonstrating that this strategy achieves a more comprehensive fusion of multi-modal data at the patch level.

\textbf{Performance Analysis:} In the experiment, night commonly used methods for object detection in remote sensing images were selected for both quantitative and qualitative comparisons. TABLE \ref{tbl:table2} lists quantitative comparison results with existing state-of-the-art methods. Although these methods have achieved a certain level of accuracy in detection, their parameter quantity and tracking speed limit their application in the field of remote sensing.
It shows that the proposed RemoteDet-Mamba achieves a mAP of 81.8\% with 71.34M parameter quantity and 24.01 Frames Per Second (FPS) detection efficiency, marking a 1.2\% improvement over the suboptimal DMM\cite{zhou2025dmm} method.

\begin{table}[ht]
    \centering
    \renewcommand\arraystretch{1.6}
    \caption{Performance comparison of different methods on the DroneVehicle dataset}
    \label{tbl:table2}
    \resizebox{\columnwidth}{!}{ % Fit to single-column width
    \begin{tabular}{c|c|ccccc|c|c|c}
        \Xhline{1pt}
         \textbf{Method} & \textbf{Modality} & \textbf{Car} & \textbf{Trunk} & \textbf{FreightCar} & \textbf{Bus} & \textbf{Van} & \textbf{mAP(\%)} & \textbf{Speed(fps)} & \textbf{Size(MB)} \\
        \Xhline{1pt}
        RetianNet\cite{lin2017focal} & \multirow{7}{*}{RGB} & 78.5 & 34.4 & 24.1 & 69.8 & 28.8 & 47.1 & 14.53 & 218 \\
        R\textsuperscript{3}Det\cite{yang2021r3det} & & 80.3 & 56.1 & 42.7 & 80.2 & 44.4 & 60.8 & - & - \\
        S\textsuperscript{2}ANet\cite{han2021align} & & 80.0 & 54.2 & 42.2 & 84.9 & 43.8 & 61.0 & - & - \\
        Faster R-CNN\cite{ren2016faster} & & 79.0 & 49.0 & 37.2 & 77.0 & 37.0 & 55.9 & 13.18 & 232 \\
        RoITransformer\cite{ding2019learning} & & 61.6 & 55.1 & 42.3 & 85.5 & 44.8 & 61.6 & 11.25 & 233 \\
        ReDet\cite{han2021redet} & & 69.48 & 47.87 & 31.46 & 77.37 & 29.03 & 51.04 & 9.11 & 125 \\
        \Xhline{1pt}
        RetianNet\cite{lin2017focal} & \multirow{7}{*}{TIR} & 88.8 & 35.4 & 39.5 & 76.5 & 32.1 & 54.5 & 14.53 & 218 \\
        R\textsuperscript{3}Det\cite{yang2021r3det} & & 89.5 & 48.3 & 16.6 & 87.1 & 39.9 & 62.3 & - & - \\
        S\textsuperscript{2}ANet\cite{han2021align} & & 89.9 & 54.5 & 55.8 & 88.9 & 48.4 & 67.5 & - & - \\
        Faster R-CNN\cite{ren2016faster} & & 89.4 & 53.5 & 48.3 & 87.0 & 42.6 & 64.2 & 13.18 & 232 \\
        RoITransformer\cite{ding2019learning} & & 89.6 & 51.0 & 53.4 & 88.9 & 44.5 & 65.5 & 11.25 & 233 \\
        ReDet\cite{han2021redet} & & 89.47 & 53.95 & 42.82 & 79.89 & 34.39 & 60.54 & 9.11 & 125 \\
        \Xhline{1pt}
        UA-CMDet\cite{sun2022drone} & \multirow{6}{*}{\makecell{RGB \\ +TIR}} & 87.51 & 60.70 & 46.80 & 87.08 & 37.95 & 64.01 & 9.12 & 234 \\
        MKD\cite{huang2023multimodal} & & 93.49 & 62.48 & 52.73 & 91.93 & 44.50 & 69.03 & \textbf{42.39} & 242 \\
        TSFADet\cite{yuan2022translation} & & 90.0 & 69.2 & 65.5 & 89.7 & 55.2 & 73.9 & 18.6 & 104.7 \\
        C\textsuperscript{2}Former\cite{yuan2024c} & & 90.2 & 68.3 & 64.4 & 89.8 & 58.5 & 74.2 & - & 118.47 \\
        DMM\cite{zhou2025dmm} & & 90.4 & 79.8 & \textbf{68.2} & 89.9 & \textbf{68.6} & 79.4 & - & 87.97 \\
        Ours & & \textbf{98.2} & \textbf{81.2} & 67.9 & \textbf{95.7} & 65.1 & \textbf{81.8} & 24.01 & \textbf{71.34} \\
        \Xhline{1pt}
    \end{tabular} 
     }
\end{table}

\begin{figure}[ht]
  \centering
  \includegraphics[width=0.35\textwidth]{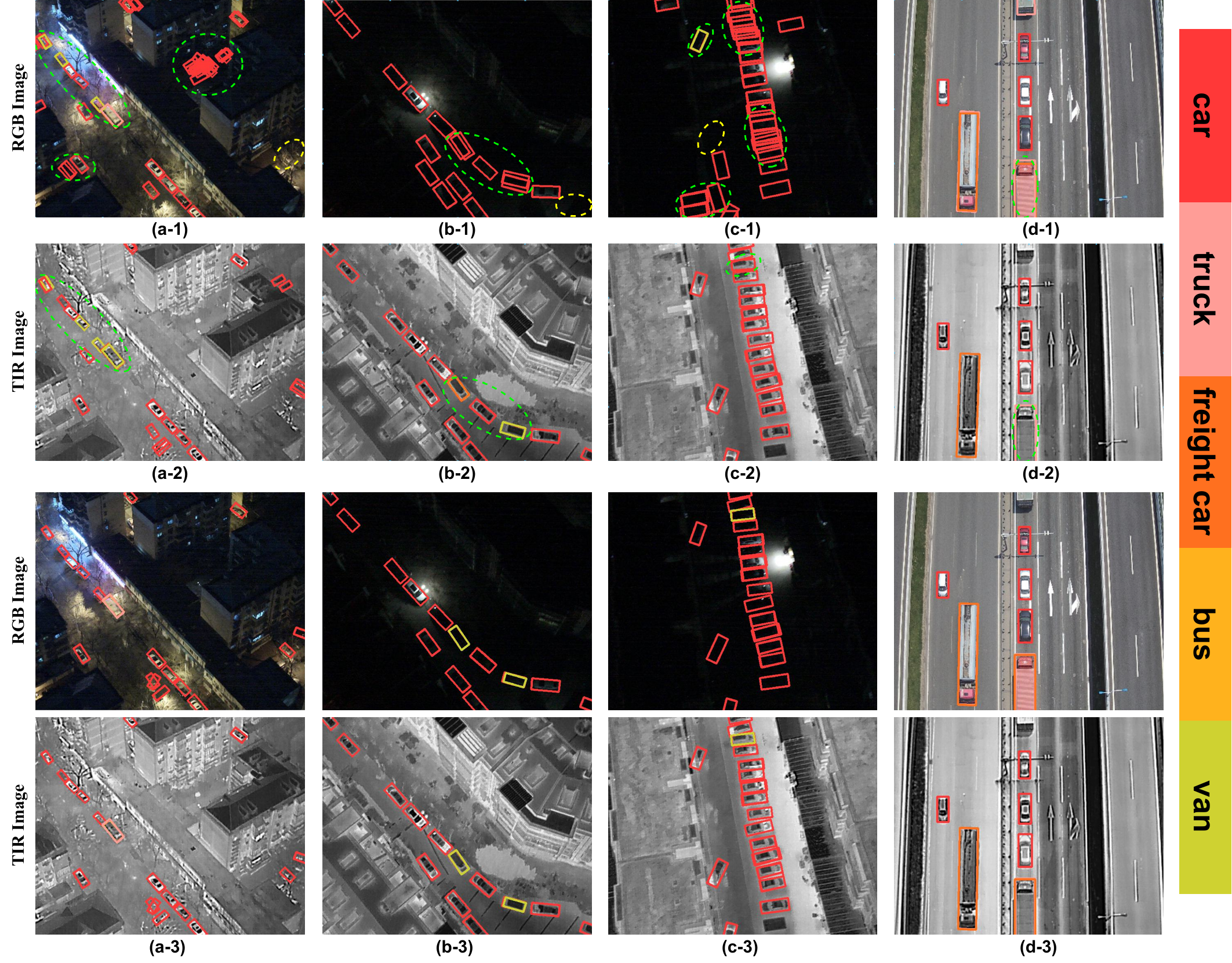}
  \caption{Detection results in (a-1)–(d-1), (a-2)–(d-2), and (a-3)–(d-3) correspond to RGB-only, TIR-only, and our RemoteDet-Mamba, respectively, on the DroneVehicle dataset.}
  \label{fig.3}
\end{figure}

Fig.\ref{fig1} shows that our model's GFLOPS scales nearly linearly with input size, demonstrating superior efficiency to both C\textsuperscript{2}Former and DMM. The quadratic complexity of Transformers causes C\textsuperscript{2}Former's memory usage to grow rapidly, restricting its maximum input size to 1092×1092 on a 24GB GPU. While DMM uses stacked Mamba layers for feature processing, our single-layer architecture achieves comparable performance with significantly lower computational costs.

Fig.\ref{fig.3} illustrates the detection results, revealing three major challenges in remote sensing object detection: small objects are easily missed (yellow boxes in the first row), densely packed objects lead to redundant bounding boxes (green boxes in the first row), and low inter-class discriminability results in misclassifications (green boxes in the second row). Under unimodal inputs (RGB or infrared, first and second rows), small objects are prone to being overlooked, dense regions generate redundant predictions, and visually similar categories are often confused due to limited lighting and texture information. In contrast, the proposed RemoteDet-Mamba (third and fourth rows) accurately detects small objects, reduces redundant bounding boxes, and mitigates class confusion, demonstrating the effectiveness of its selective fusion mechanism in addressing the challenges of small object size, dense distribution, and low inter-class discriminability.

\section{Conclusions}
\label{sec:conclu}

In conclusion, this paper presents RemoteDet-Mamba, a novel lightweight multimodal UAV object detection framework based on a hybrid architecture that combines CNNs and the Mamba. This design leverages CNNs’ ability to capture multi-scale local features and Mamba’s linear complexity with global receptive fields. To enable efficient fusion, we introduce a single-layer Mamba-based CFM module with four-directional patch-level scanning, which effectively decouples dense targets and enhances small object detection. The proposed framework achieves high accuracy with low computational overhead, making it well-suited for real-time remote sensing applications.

\bibliographystyle{IEEEbib}
\bibliography{strings,refs}

\end{document}